\documentclass[runningheads]{llncs}

\usepackage{graphicx}
\usepackage[inline]{enumitem} 
\usepackage{float}
\usepackage{subcaption}
\usepackage{algorithm2e}
\usepackage{tabularx}

\usepackage{tikz}
\usetikzlibrary{shapes.geometric, arrows, spy}
\tikzstyle{process} = [rectangle, minimum width=4.3cm, minimum height=0.7cm, text centered, draw=black]
\tikzstyle{arrow} = [thick,->,>=stealth]
\definecolor{uibkblue}{cmyk}{1,0.6,0,0.65}
\definecolor{uibkgrayll}{cmyk}{0,0,0,0.1}
\definecolor{uibkorangel}{rgb}{1.00,0.90,0.76}

\begin{document}
\newcommand{\antonio}[1]{\emph{\textcolor{red}{(Antonio: #1)}}}
\newcommand{\Josef}[1]{\emph{\textcolor{blue}{(Josef: #1)}}}
\newcommand{\david}[1]{\emph{\textcolor{orange}{(David: #1)}}}

%
% TITLE
%
\title{Training Deep Capsule Networks with Residual Connections}
\titlerunning{Deep Residual Capsule Networks}

\author{Josef Gugglberger\inst{1} \and
David Peer\inst{1}\orcidID{0000-0001-9028-0920} \and \\
Antonio Rodr\'iguez-S\'anchez\inst{1}\orcidID{0000-0002-3264-5060}}
\authorrunning{J. Gugglberger et al.}
\institute{University of Innsbruck, Austria\\
    \email{josef.gugglberger@student.uibk.ac.at},
    \email{david.peer@outlook.com},
    \email{antonio.rodriguez-sanchez@uibk.ac.at} 
}
\maketitle

%
% ABSTRACT
%
\begin{abstract}

Capsule networks are a type of neural network that have recently gained increased popularity. They consist of groups of neurons, called capsules, which encode properties of objects or object parts. The connections between capsules encrypt part-whole relationships between objects through routing algorithms which route the output of capsules from lower level layers to upper level layers. Capsule networks can reach state-of-the-art results on many challenging computer vision tasks, such as MNIST, Fashion-MNIST and Small-NORB. However, most capsule network implementations use two to three capsule layers, which limits their applicability as expressivity grows exponentially with depth \cite{expressive_power}. One  approach  to overcome such limitation would be to train deeper network architectures, as it has been done for convolutional neural networks with much increased success. In this paper we propose a methodology to train deeper capsule networks using residual connections, which is evaluated  on  four  datasets  and  three  different  routing algorithms. Our experimental results show that in fact, performance increases when training deeper capsule networks. The source code is available on

\url{https://github.com/moejoe95/res-capsnet}. 

\end{abstract}

%
% INTRO
%
\section{Introduction}

Capsule Networks were introduced by Sabour et al. \cite{sabour2017dynamic}, although the idea behind  \textit{capsules} was earlier introduced by Hinton et al. \cite{hinton2011transforming}. A capsule represents a group of neurons, and each neuron in a capsule can be seen as an instantiation parameter of some object in the image. In other words, a capsule is a vector of neurons, where its length would define the capsule's \textit{activation}, representing the presence of an object or object-part in the input. The vector's orientation relates to the properties (position, pose, size, etc.) of the object. Capsules can also be in matrix form, together with a scalar variable that represents its activation, as it was described by Hinton et al. in \cite{hinton2018matrix} as \textit{Matrix Capsules}.

Capsules of a lower layer \textit{vote} for the pose of capsules in the upper layer, by multiplying its own pose with a transformation matrix. This transformation matrix contains learnable parameters, which represents part-whole relationships of real objects. Transformation matrices are obtained through training and model viewpoint-invariant part-whole relationships, so that the change in viewpoint to an object does not change the agreement between capsules. This makes capsule networks better suited than classical CNNs for 3D viewpoint object recognition \cite{hinton2018matrix}. 

The aforementioned votes are weighted by a coefficient, which is computed dynamically by a \textit{routing} algorithm. This routing algorithm computes the agreement between two capsules, where a high value is given to strongly agreeing capsules. The first dynamic routing algorithm - called routing-by-agreement or RBA- was proposed by Sabour et al. \cite{sabour2017dynamic}. RBA computes the agreement between capsules by a dot product of the predicted activation of the lower level capsule with the pose of the current capsule. Later, Hinton et al. \cite{hinton2018matrix} published an improved routing algorithm based on expectation maximization, called EM routing. 
Routing capsule networks with RBA (\cite{sabour2017dynamic}) improved the state-of-the-art accuracy on MNIST and outperformed previous approaches on an MNIST-like dataset with highly overlapping digits. Through EM routing, Hinton et al. (\cite{hinton2018matrix}) reported new state-of-the-art performances on the Small-NORB dataset, reducing the error rate by $45 \%$. 

In order to obtain the best performance results, it is of great importance to design very deep networks as has been shown with classical CNNs \cite{simonyan2014very}. However, most capsule network implementations use two to three capsule layers, which may be the reason behind their low performance on more complex data when compared to CNNs. We believe that the performance of capsule networks can be also greatly increased by designing and training deeper network architectures. CNNs use skip connections \cite{he2015} to stabilize the training process but have not yet been considered for the training of deep capsule networks. 
Rajasegaran \cite{rajasegaran2019deepcaps} et al. uses residual connections, but only one routing iteration for all capsule layers except the last one. They claim that a capsule layer with only one routing iteration can be approximated with a classical single 2D convolutional layer and therefore only one capsule layer exists in their architecture \footnote{\url{https://github.com/brjathu/deepcaps/issues/15}}. In contrast we show in this paper how skip connections can be used between multiple capsule layers and not only convolutional layers.

In this work, we will show experimentally that the performance of deep capsule networks can be improved by adding residual connections between capsule layers for three different and commonly used routing algorithms: routing-by-agreement (RBA) by Sabour et al. \cite{sabour2017dynamic}, EM routing by Hinton et al. \cite{hinton2018matrix}, and scaled-distance-agreement (SDA) routing by Peer et al. \cite{peer2018increasing}. The evaluation was performed on four well-known datasets: MNIST, Fashion-MNIST, SVHN and Small-NORB. 

%
% RELATED WORK
%
\section{Related Work}

The first capsule network with a dynamic routing mechanism was proposed by Sabour et al. \cite{sabour2017dynamic}. The authors showcased the potential of capsule networks on MNIST, using an algorithm called \textit{routing-by-agreement}. Later, Hinton et al. \cite{hinton2018matrix} came up with the more powerful EM routing algorithm. Since then, various different routing algorithms have been presented presented, every new implementation obtains better results when compared to previous methods.
Other routing algorithms worth of mention are \textit{scaled distance agreement routing} \cite{peer2018increasing}, \textit{inverted dot-product attention routing}  \cite{tsai2020capsules} and \textit{routing via variational bayes} \cite{ribeiro2020capsule} just to name a few. 

Capsule networks show state-of-the-art performance on many simple datasets like MNIST as well as on datasets where we want to model viewpoint invariant part-whole relationships, like Small-NORB. 
However, the design and training of capsule networks for more complex data is still an open question. Capsule networks do not perform at the same level as modern CNN architecture. For example, in CIFAR-10 (\cite{krizhevsky2009learning}), capsule networks reach an error rate of about $9 \%$ (\cite{ribeiro2020capsule}), compare it to the $3 \%$ error rate of CNN approaches \cite{wistuba2019survey}.

Late implementations have tried to improve capsule networks results on CIFAR-10. Xi et al. \cite{xi2017capsule} were able to improve the baseline model by using a more powerful feature extractor in front of the capsule network, and training the network in an 4-model ensemble. Similarly, Ai et al. \cite{ai2021rescaps} presented a capsule network named \textit{ResCaps}, having a residual sub-network in front of the capsule network. Although called ResCaps, the authors do not use skip connections between capsule layers, as opposed to our work.
Rajasegaran et al. \cite{rajasegaran2019deepcaps} were able to reach remarkable results on CIFAR-10, Fashion-MNIST and SVHN by stacking up 16 convolutional capsule layers with residual connections. They report an accuracy of $92.74\%$ on CIFAR-10, $97.56\%$ on Fashion-MNIST, and $94.73\%$ on SVHN. Differently to the work presented in this paper, the authors did not use dynamic routing in the network, but only on one layer of a residual connection in the last block. Peer et al. \cite{peer2019limitations} proposed a way to train deeper capsule networks that use routing-by-agreement or EM routing. The authors proved theoretically and showed experimentally, that these two routing algorithms can be improved by adding a bias term to the pre-activations in RBA and adding a bias term to the pose matrix in EM routing. 

On the other hand, training very deep convolutional neural networks shows excellent results for a wide range of tasks in computer vision. Simonyan et al. \cite{simonyan2014very} conducted a comprehensive study on how network depth influences the performance of CNNs. In that work, VGG provided an error rate of $23.7 \%$ on ImageNet \cite{deng2009imagenet}, while the older and not so deep AlexNet \cite{krizhevsky2012imagenet} performed with an error rate of $40 \%$. At some point, simply stacking up a higher number of layers does not  improve the accuracy of a network, because of the \textit{vanishing/exploding} gradient problem \cite{hochreiter1991untersuchungen}. Proper weight initialization, as done by Glorot et al.  \cite{glorot2010understanding}, and normalization techniques, such as batch normalization \cite{ioffe2015batch} where able to overcome those issues to some extend. Even so, very deep networks still face a \textit{degradation} problem \cite{he2015}, such that when the the accuracy gets saturated, it drops fast, resulting in a high training error. The solution to this degradation problem was presented as the \textit{deep residual learning framework}. Additionally, the \textit{conflicting bundles problem} \cite{peer2021conflicting} becomes more present using deeper networks, where the floating point precision of GPUs and CPUs can be another reason for this problem. The authors showed in that work that  residual connections can help to resolve this problem of underperformance in the case very deep convolutional neural networks.

%
% DEEP CAPSULE NETWORKS
%
\section{Training Deep Capsule Networks}

The depth of a CNN - probably, the most successful network architecture in vision - is a very important hyper-parameter because expressivity has been proved to grow exponentially with depth \cite{expressive_power}. We will show in this chapter how to train deep capsule networks in order to improve the performance of this type of neural networks. Simply "stacking up" layers is not a proper strategy to follow and leads to failure when using routing algorithms such as routing-by-agreement and EM routing \cite{xi2017capsule}. We hypothesize we can be succeed at training deep capsules through the use of residual connections between capsule layers. In this section we will analyze in detail this hypothesis and provide the details of the network architecture for the different routing procedures.

\subsection{Routing Algorithms}
Dynamic routing in capsule layers is a much more sophisticated strategy than just the classical pooling operation present in CNNs. We conducted our experiments using the following three routing algorithms, as they are widely used:
\begin{enumerate*}
    \item Routing-by-agreement (RBA) \cite{sabour2017dynamic},
    \item Expectation-maximization (EM) routing \cite{hinton2018matrix}, and
    \item Scaled-distance-agreement (SDA) routing \cite{peer2018increasing}.
\end{enumerate*}

The goal of a routing algorithm is to compute the agreement between lower level capsules (child capsules) and higher level capsules (parent capsules). This agreement is represented by the length of a vector and acts as a weight, determining the parent(s) capsule(s) to which a child capsule \emph{routes} its output. Routing is typically implemented iteratively, requiring usually just two to five iterations for convergence \cite{sabour2017dynamic,hinton2018matrix,peer2018increasing,ribeiro2020capsule}. We use two routing iterations for all our experiments because we found that this achieves high accuracy.

RBA and SDA-routing (Algorithm \ref{alg:sda_routing}) have a very similar structure. The iterative process starts by calculating the coupling coefficients between child capsules and parent capsules, which sum up to 1 by applying the \textit{softmax} function over the agreements. Each capsule in the higher level layer then calculates a weighted average with the coupling coefficient and the votes from the lower level capsule layer, which are \textit{squashed} to a value between $0$ and $1$. In RBA, the agreement is computed by a dot product of the current pose and the votes from each child capsule. Finally the agreement tensor is updated and the procedure continues for a fixed amount of routing iterations. 
On the other hand, SDA calculates the agreement by inverse distances instead of the dot product, as can be seen in line $8$ of algorithm \ref{alg:sda_routing}. This ensures that active lower level capsules do not couple with inactive higher level capsules and that the prediction is limited by the activation of the corresponding capsule (line 2).

\begin{algorithm}[t][1]
\LinesNumbered
\KwIn{$v_i$, $\hat{u}_{j|i}$, $r$, $l$}
 $b_{ij} \leftarrow 0$ \\
 $\hat{u}_{j|i} \leftarrow min(||v_i||, ||\hat{u}_{j|i}||) \cdot \frac{\hat{u}_{j|i}}{||\hat{u}_{j|i}||} $\\
 \For{$r$ iterations}{
  $c_{ij} \leftarrow \frac{b_{ij}}{\sum_k exp(b_{ik})} $ \\
  $s_{j} \leftarrow \sum_i c_{ij} \cdot \hat{u}_{j|i}$ \\
  $v_{j} \leftarrow \frac{||s_j||^2}{1+||s_j||^2} \cdot \frac{s_j}{||s_j||} $ \\
  $t_i \leftarrow \frac{log(0.9 (J-1)) - log(1-0.9)}{-0.5 mean^J_j(|| \hat{u}_{j|i} - v_j||)}$ \\
  $b_{ij} \leftarrow || \hat{u}_{j|i} - v_j || \cdot t_i $
 }
 \Return $v_{j}$
 \caption{Scaled distance agreement routing algorithm. $\forall$ capsules $i$ of child capsule layer $l$ and $J$ parent capsules on layer $l+1$, with $r$ routing iterations and predictions $\hat{u}_{j|i}$ from child capsule with activation $v_i$.}
 \label{alg:sda_routing}
\end{algorithm}

EM routing works by fitting the mixtures of Gaussians parameters through expectation maximization. This Gaussian mixture model clusters datapoints into Gaussian distributions, each described by a mean $\mu$ and a standard deviation $\sigma$. Starting with random assignments, the expectation maximization (EM) algorithm iteratively assigns the datapoints to Gaussians and recomputes $\mu$ and $\sigma$. The EM routing algorithms formulates the problem of routing as a clustering problem, assigning lower level capsules to higher level capsules. EM routing was proposed first together with Matrix-Capsules, which uses a matrix and a separate scalar for representing the activation instead of the length of the vector.

\subsection{Applying residual learning to capsule networks}

The starting point of residual learning is that a deep network should perform at least as good as a shallower one \cite{he2015}. In order to make this possible, identity-shortcut connections are inserted, skipping one or more layers and simply adding their output to the output of a deeper layer. Those skip connections do not contain learnable parameters, and as such, layers can be replaced by the identity function. In deep neural networks performance decreasing layers occur with higher probability. It is known that residual connections bypass those layers such that the accuracy of the trained model is increased \cite{peer2021conflicting}.

\begin{figure}[t!]
     \centering
     \begin{subfigure}{0.40\textwidth}
        \resizebox{0.7\textwidth}{6.3cm}{
        \begin{tikzpicture}[node distance=1.1cm]
            % layers of network
            \node (input) [] {\textbf{Input}};
            \node (conv1) [process, below of=input] {\textbf{Conv} + \textbf{ReLu}};
            \node (primary) [process, below of=conv1] {\textbf{PrimaryCapsules}};
            \node (capsule) [fill=uibkgrayll, process, below of=primary] {\textbf{Capsule Net}};
            \node (class) [process, below of=capsule] {\textbf{ClassCapsules}};
            \node (recon) [fill=uibkgrayll, draw, rectangle, below of=class, minimum height=0.7cm, xshift=0.725cm] {\textbf{Reconstruction Net}};
            \node (norm) [fill=uibkorangel, circle, below of=recon, xshift=-1.75cm] {\textbf{Norm}};
            \node (out1) [below of=norm] {\textbf{Class. Output}};
            \node (out2) [right of=out1, xshift=1.5cm] {\textbf{Recon. Output}};
        
            % arrows with dimensions
            \draw [arrow] (input) {} -- (conv1) {};
            \draw [arrow] (conv1) node[below, xshift=1.8cm, yshift=-0.35cm] {} -- (primary) {};
            \draw [arrow] (primary) node[below, xshift=1.8cm, yshift=-0.35cm] {} -- (capsule) {};
            \draw [arrow] (capsule) node[below, xshift=1.8cm, yshift=-0.35cm] {} -- (class) {};
            \draw [arrow] (class) node[below, xshift=1.8cm, yshift=-0.35cm] {} -- (recon) {};
            \draw [arrow] (class) [bend angle=45, bend right] to  (norm) {};
            \draw [arrow] (recon) [bend left] to (out2) {};
            \draw [arrow] (norm) [bend right] {} -- (out1) {};
        \end{tikzpicture}
        }
        \caption{High level overview of our capsule network. The \textit{Capsule Net} block represents a sub-network of capsule layers, which contains from $1$ to $15$ capsule layers, arranged as shown in Figure \ref{fig:skip}.}
        \label{fig:arch}
     \end{subfigure}
     \hfill
     \begin{subfigure}{0.45\textwidth}
        \resizebox{0.6\textwidth}{4cm}{
        \begin{tikzpicture}[node distance=1.1cm]
            % layers of network
            \node (input) {\textbf{Input}};
            \node (capsule1) [fill=uibkgrayll, draw, rectangle, below of=input, minimum height=0.5cm] {\textbf{Capsule Layer}};
            \node (capsule2) [fill=uibkgrayll, draw, rectangle, below of=capsule1, minimum height=0.5cm] {\textbf{Capsule Layer}};
            \node (add) [fill=uibkorangel, circle, below of=capsule2] {\textbf{Add}};
            \node (output) [below of=add] {\textbf{Output}};
        
            \draw [arrow] (input) {} -- (capsule1) {};
            \draw [arrow] (input) [bend angle=90, bend right] to (add) {};
            \draw [arrow] (capsule1) node[below, xshift=1cm, yshift=-0.35cm] {} -- (capsule2) {};
            \draw [arrow] (capsule2) node[below, xshift=1cm, yshift=-0.35cm] {} -- (add) {};
            \draw [arrow] (add) {} -- (output) {};
        \end{tikzpicture}
        }
        \vspace{1.2cm}
        \caption{Skip connections are element-wise added to the output of the two layer above capsule layer.}
        \label{fig:skip}
    \end{subfigure}
    \vspace{-0.25cm}
    \caption{Architecture of our capsule network (Figure \ref{fig:arch}) with detailed view of the capsule sub-network (Figure \ref{fig:skip}). Given dimensions apply to the datasets MNIST and Fashion-MNIST.}
\end{figure}
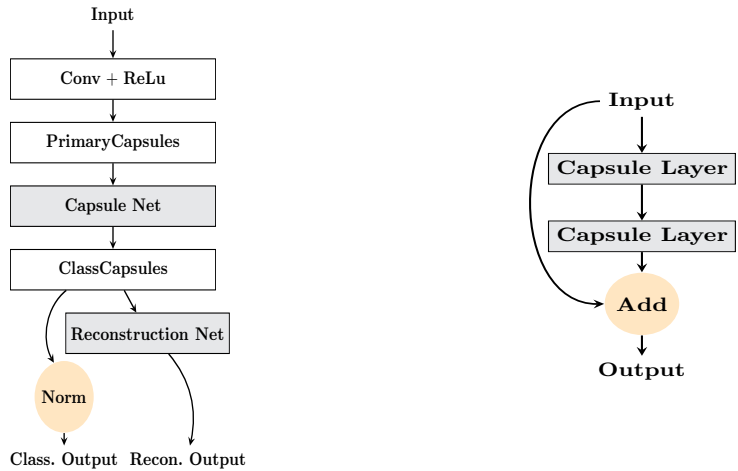

We include shortcut connections between capsule layers as explained next. The output of a capsule layer is element-wise added to the capsule layer that is located two layers deeper into the network. We add the shortcut connection after routing has happened, because this approach yielded the best results. We also did experiments on adding skip connections before routing or doing another squashing after the element-wise addition but did not achieve good results. This design is shown in figure \ref{fig:skip} and is in stark difference to \emph{ResCaps}, presented by Ai et al. \cite{ai2021rescaps}, where a residual sub-network is located in front of the capsule network. The authors replaced the single convolutional layer that we used before the \emph{PrimaryCapsule} layer by a residual sub-network, to provide better features to the capsule network. The implementation of Rajasegaran \cite{rajasegaran2019deepcaps} et al. uses only one routing iteration for all capsule layers except the last one i.e. all layers are implemented as classical 2d-convolutional layers except the last one, which is also confirmed by the original authors on GitHub\footnote{\url{https://github.com/brjathu/deepcaps/issues/15}}. Opposed to that, our network uses more than one routing iteration in every capsule layer. Additionally, we will show in the experimental evaluation, that the performance will not drop as the network depth is increased (also known as the degradation problem), regardless of the used routing algorithm.

%
% EXPERIMENTS
%
\section{Experimental evaluation}

In this section we describe the experimental setup. First, we will explain in detail the network architecture and setup. Next, we will 
we will describe the datasets used to evaluate our model. We will finish this section with the results of our experiments.

\subsection{Setup}
The general architecture is shown in figure \ref{fig:arch}. The input is fed into a normal convolutional layer with a fixed kernel size of $(9,9)$ and a stride of 1. We then apply another convolution with a a $(9,9)$ kernel and a stride of 2 inside the layer \textit{PrimaryCapsules} and reshape into capsule form. Both convolutions use the ReLu activation function. The first \textit{Capsule} layer has 512 incoming capsules and 32 outgoing capsules with a dimension of 8. Afterwards, we feed into a sub-network containing capsule layers, which contains $1$ to $15$ fully connected capsule layers, where each capsule is represented by a 12 element vector. The last capsule layer, the \textit{ClassCapsule} layer, has one capsule of dimension 16 for each class contained in the dataset. Dynamic routing between capsule layers is either done by RBA, SDA or EM routing.

Residual connections were added between every second layer in the \textit{Capsule Net} sub-network of Figure \ref{fig:arch} as shown in Figure \ref{fig:skip}. Skip connections do not contain any learnable parameters since the dimensionality of capsule layers do not vary across the inserted connections. Figure \ref{fig:skip} shows that the tensors coming from the skip connections are element-wise added to the output of the two layer deeper capsule layer.

The reconstruction network contains three densely connected layers, with $512$, $1024$ and $576$ neurons. Layer one and two use the \textit{ReLu} activation, while the last layer implements a \textit{sigmoid} activation. The reconstruction network is used to compute the \textit{reconstruction loss} (\cite{sabour2017dynamic}), used for training the capsule network. The sum of squared differences between the input pixels of the image and the output from the reconstruction network is used in the objective function. The overall loss function used for training is the sum of the \textit{margin} loss and the reconstruction loss, which is weighted by a scalar factor. The margin loss (\cite{sabour2017dynamic}) indicates that the length of a capsule should be large if and only if the class is present in the input image.

Weights of the transformation matrices were initialized randomly from a normal distribution with a standard deviation of $0.2$ and a mean of $0$. The weights of the bias terms were initialized with a constant value of $0.1$. While training, the network receives random crops of 24x24, and on inference center crops are used. We trained with a batch size of $128$, and used Adam optimizer with a constant learning rate of $1^{-4}$. For networks deeper than $13$ layers we used a batch size of $64$, because of memory constraints of our GPU. The network was trained with a combined objective function of margin loss and reconstruction loss. The latter was added with a weight of $1^{-5}$. Each model was trained for 30 epochs,  with and without the use of shortcut connections for comparison.

Our implementation\footnote{https://github.com/moejoe95/res-capsnet} of the capsule network uses Tensorflow \cite{tensorflow2015-whitepaper} Version 2.3.

\subsection{Datasets \& Data Augmentation}

We evaluated our model on four different datasets:  MNIST \cite{MNIST}, Fashion-MNIST \cite{fashionMNIST}, Small-NORB \cite{smallNORB}, and SVHN \cite{SVHN}.

MNIST is a dataset of 28x28 greyscale images of handwritten digits, containing 10 classes, representing the digits from 0-9. Fashion-MNIST is a little bit more involved than MNIST, but with a very similar structure. It also contains greyscale images of size 28x28 and there are 10 distinct classes, each representing a type of clothing. Small-NORB is a dataset of greyscale 96x96 images showing objects from different elevations, azimuths and under different lighting conditions. It contains 5 classes of toys. SVHN is a dataset of 32x32 RGB images showing real world pictures of house numbers. As MNIST, it contains 10 classes representing the digits from 0-9.

We applied data augmentation by adding random brightness with  intensities $[-0.25, 0.25)$ to the images in Small-NORB and SVHN. On Fashion-MNIST we augmented the data by horizontally flipping pixels of images with a probability of $50 \%$. After data augmentation, we normalized per image to have zero mean and a variance of 1. We scaled down images of Small-NORB to the size $28x28$. 

\begin{table}[t]
\caption{Test accuracy for capsule networks using RBA with 3-11 capsule layers layers, trained with and without skip connections.}
\noindent\makebox[\textwidth]{%
    \begin{tabular}{| l l l l l l l l l l l |}
    \hline
    \textbf{Dataset} & \textbf{Method} & \textbf{3} & \textbf{4} & \textbf{5} & \textbf{6} & \textbf{7} & \textbf{8} & \textbf{9} & \textbf{10} & \textbf{11} \\ \hline
    MNIST & RBA & 0.995 & 0.993 & 0.990 & 0.889 & 0.114 & 0.114 & 0.114 & 0.114 & 0.114 \\
     & RBA+Skip & 0.995 & \textbf{0.995} & \textbf{0.995}& \textbf{0.993} & \textbf{0.989} & \textbf{0.993} & \textbf{0.992} & \textbf{0.99} & \textbf{0.982} \\ \hline
    Fashion & RBA & \textbf{0.892} & 0.880 & 0.868 & 0.752 & 0.100 & 0.100 & 0.100 & 0.100 & 0.100 \\
    MNIST & RBA+Skip &  0.886 & \textbf{0.891} & \textbf{0.891} & \textbf{0.883} & \textbf{0.870} & \textbf{0.882} & \textbf{0.882} & \textbf{0.870} & \textbf{0.841} \\ \hline
    SVHN & RBA & \textbf{0.926} & 0.912 & 0.860 & 0.613 & 0.410 & 0.196 & 0.196 & 0.196 & 0.196 \\
     & RBA+Skip & 0.923 & \textbf{0.928} & \textbf{0.921} & \textbf{0.916} & \textbf{0.865} & \textbf{0.905} & \textbf{0.909} & \textbf{0.858} & \textbf{0.791} \\ \hline
    Small & RBA & 0.889 & 0.859 & 0.781 & 0.531 & 0.200 & 0.200 & 0.200 & 0.200 & 0.200 \\
    NORB & RBA+Skip & \textbf{0.891} & \textbf{0.885} & \textbf{0.883} & \textbf{0.853} & \textbf{0.822} & \textbf{0.859} & \textbf{0.870} & \textbf{0.852} & \textbf{0.522} \\
     \hline
    \end{tabular}
}
\label{tab:rba}
\end{table}

\begin{table}[t]
\caption{Test accuracy for capsule networks using SDA routing with 3-16 capsule layers, trained with and without skip connections.1}
\noindent\makebox[\textwidth]{%
    \begin{tabular}{| l l l l l l l l l l l l l l l l|}
    \hline
    \textbf{Dataset} & \textbf{Method} & \textbf{3} & \textbf{4} & \textbf{5} & \textbf{6} & \textbf{7} & \textbf{8} & \textbf{9} & \textbf{10} & \textbf{11} & \textbf{12} & \textbf{13} & \textbf{14} & \textbf{15} & \textbf{16} \\ \hline
    MNIST & SDA & 0.994 & 0.994 & 0.994 & 0.993 & 0.992 & 0.993 & 0.993 & 0.991 & 0.991 & 0.986 & 0.972 & 0.972 & 0.864 & 0.114 \\
     & SDA+Skip & 0.994 & 0.994 & 0.994 & \textbf{0.994} & \textbf{0.994} & \textbf{0.994} & \textbf{0.993} & \textbf{0.993} & \textbf{0.993} & \textbf{0.992} & \textbf{0.993} & \textbf{0.992} & \textbf{0.991} & \textbf{0.990} \\ \hline
    Fashion & SDA & 0.889 & \textbf{0.893} & 0.887 & \textbf{0.890} & 0.881 & 0.877 & 0.883 & 0.876 & 0.870 & 0.874 & 0.819 & 0.747 & 0.715 & 0.100 \\ 
    MNIST & SDA+Skip & \textbf{0.892} & 0.889 & \textbf{0.890} & 0.888 & \textbf{0.890} & \textbf{0.888} & \textbf{0.887} & \textbf{0.883} & \textbf{0.884} & \textbf{0.884} & \textbf{0.886} & \textbf{0.884} & \textbf{0.882} & \textbf{0.883} \\ \hline
    SVHN & SDA & \textbf{0.922} & 0.916 & 0.913 & 0.902 & 0.899 & 0.895 & 0.896 & 0.893 & 0.863 & 0.671 & 0.784 & 0.595 & 0.631 & 0.323 \\ 
     & SDA+Skip & 0.919 & \textbf{0.924} & \textbf{0.920} & \textbf{0.913} & \textbf{0.910} & \textbf{0.910} & \textbf{0.910} & \textbf{0.904} & \textbf{0.905} & \textbf{0.900} & \textbf{0.902} & \textbf{0.898} & \textbf{0.901} & \textbf{0.896} \\ \hline
    Small & SDA & 0.899 & 0.893 & 0.873 & 0.878 & 0.876 & 0.871 & 0.876 & 0.860 & 0.857 & 0.865 & 0.714 & 0.830 & 0.523 & 0.383 \\ 
    NORB & SDA+Skip & \textbf{0.903} & \textbf{0.910} & \textbf{0.897} & \textbf{0.881} & \textbf{0.892} & \textbf{0.881} & \textbf{0.878} & \textbf{0.871} & \textbf{0.867} & \textbf{0.876} & \textbf{0.864} & \textbf{0.886} & \textbf{0.872} & \textbf{0.855} \\
     \hline
    \end{tabular}}
\label{tab:sda}
\end{table}

\begin{table}[t]
\caption{Test accuracy for capsule networks using EM routing with 3-16 capsule layers, trained with and without skip connections.}
\noindent\makebox[\textwidth]{%
    \begin{tabular}{| l l l l l l l l l l l l l l l l|}
    \hline
    \textbf{Dataset} & \textbf{Method} & \textbf{3} & \textbf{4} & \textbf{5} & \textbf{6} & \textbf{7} & \textbf{8} & \textbf{9} & \textbf{10} & \textbf{11} & \textbf{12} & \textbf{13} & \textbf{14} & \textbf{15} & \textbf{16}\\ \hline
    MNIST & EM & 0.995 & 0.995 & 0.995 & 0.994 & 0.994 & 0.994 & 0.994 & 0.993 & 0.992 & 0.992 & 0.989 & 0.988 & 0.496 & 0.984 \\
 & EM+Skip & 0.995 & 0.995 & 0.995 & 0.994 & 0.994 & 0.994 & 0.994 & 0.993 & 0.992 & 0.992 & \textbf{0.992} & \textbf{0.992} & \textbf{0.991} & \textbf{0.988} \\ \hline
    Fashion & EM & 0.890 & 0.890 & 0.882 & 0.881 & 0.879 & 0.874 & 0.875 & 0.875 & 0.869 & 0.871 & 0.867 & 0.867 & 0.856 & 0.859 \\
    MNIST & EM+Skip & \textbf{0.892} & 0.890 & \textbf{0.888} & \textbf{0.886} & \textbf{0.880} & \textbf{0.881} & \textbf{0.880} & \textbf{0.879} & \textbf{0.882} & \textbf{0.875} & \textbf{0.880} & \textbf{0.873} & \textbf{0.879} & \textbf{0.870} \\ \hline
    SVHN & EM & 0.923 & 0.911 & 0.897 & 0.897 & 0.883 & 0.887 & 0.883 & 0.880 & 0.878 & 0.873 & \textbf{0.871} & 0.864 & \textbf{0.847} & 0.797 \\
   & EM+Skip & \textbf{0.931} & \textbf{0.930} & \textbf{0.919} & \textbf{0.906} & \textbf{0.906} & \textbf{0.902} & \textbf{0.890} & \textbf{0.890} & \textbf{0.894} & \textbf{0.889} & 0.480 & \textbf{0.880} & 0.414 & \textbf{0.871} \\ \hline
    Small & EM & \textbf{0.904} & 0.894 & 0.866 & 0.880 & \textbf{0.873} & \textbf{0.872} & \textbf{0.873} & 0.857 & \textbf{0.872} & 0.846 & 0.846 & 0.849 & 0.836 & 0.827 \\
    NORB & EM+Skip & 0.896 & \textbf{0.898} & \textbf{0.888} & \textbf{0.881} & 0.872 & 0.863 & 0.872 & \textbf{0.879} & 0.867 & \textbf{0.851} & \textbf{0.871} & \textbf{0.864} & \textbf{0.868} & \textbf{0.850} \\
     \hline
    \end{tabular}
}
\label{tab:em}
\end{table}

\subsection{Results}

Tables \ref{tab:rba} and \ref{tab:sda} show the results of training capsule networks using the two similar routing algorithms RBA and SDA, comparing the cases where there were no skip connections with the ones in the presence of skip connections for MNIST, Fashion-MNIST, SVHN and Small-NORB. We can see that both routing algorithms benefit from skip connections. These results also show that SDA is more robust than RBA for routing deeper capsule networks with and without skip connections. In the case of no skip connections, RBA-routing shows a dramatic decrease in performance after 7 layers, while SDA holds up up to depths of 13 layers. In some cases, RBA barely performs better than chance after just using 7 layers, such as in MNIST, Fashion-MNIST and Small-NORB. On the other hand, if we use skip connections, RBA performs well up to 11 layers of depth and SDA up to at least 16 layers. Table \ref{tab:em} provides the results for capsule networks using EM-routing for the four used datasets. This routing algorithm exhibits the highest robustness of the three used in this work. While there is also a benefit of using skip connections, the increase of performance when using EM routing is smaller as with RBA and SDA.

\begin{figure}[!ht]
% RBA
\noindent\makebox[\textwidth]{%
\begin{subfigure}{.35\textwidth}
  \centering
  \includegraphics[width=\linewidth]{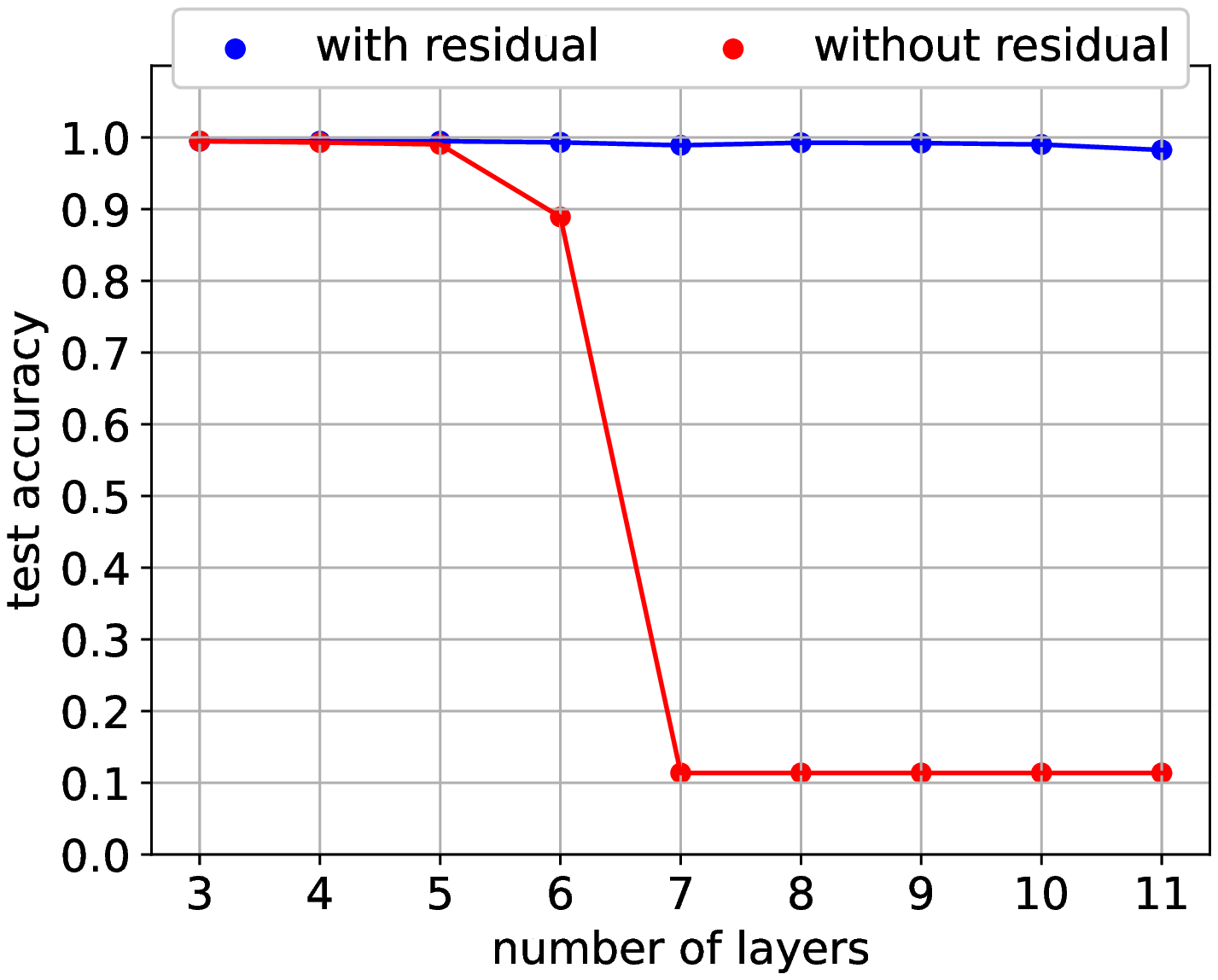}
  \caption{}
  \label{fig:mnist_rba}
\end{subfigure}
\begin{subfigure}{.35\textwidth}
  \centering
  \includegraphics[width=\linewidth]{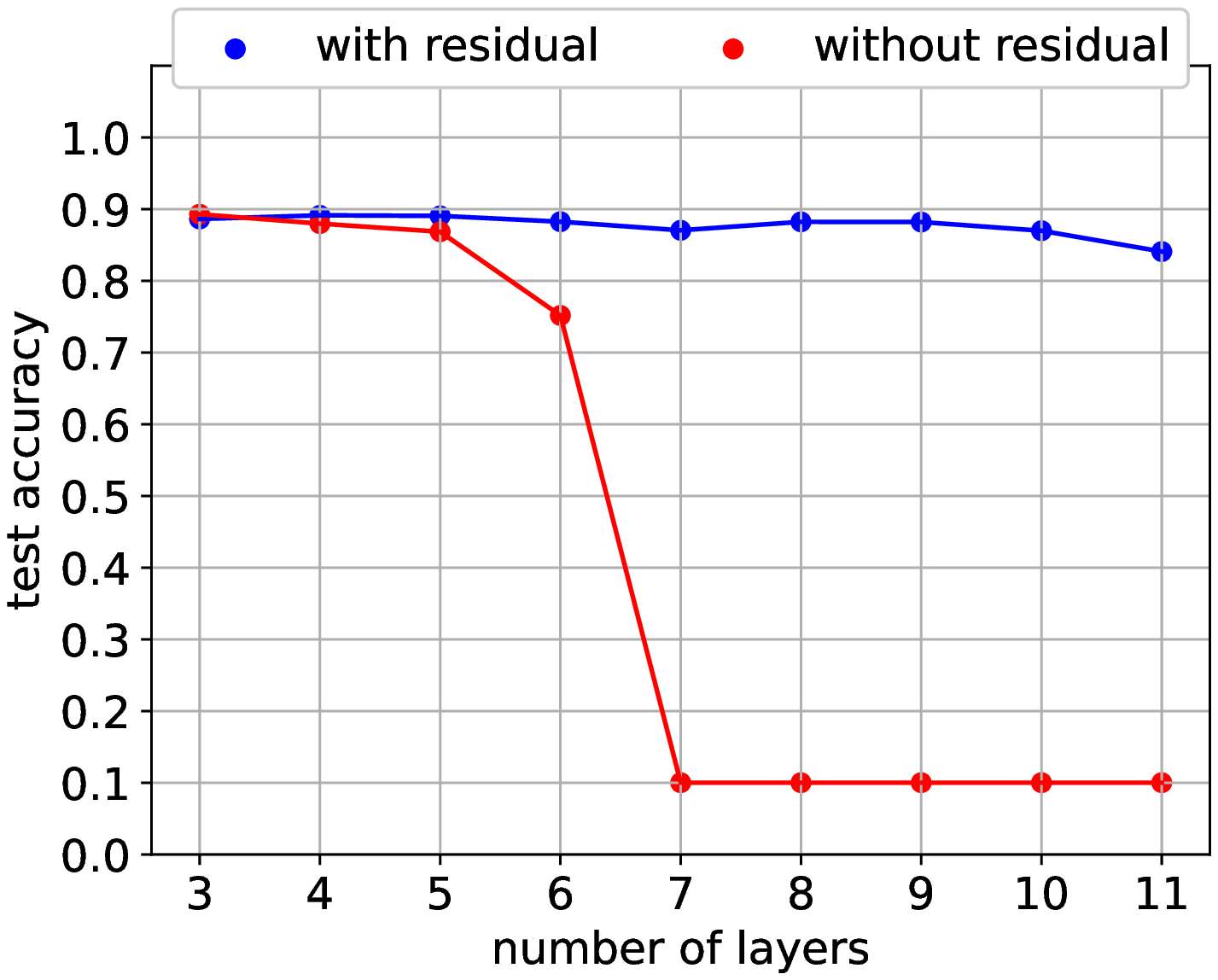}
  \caption{}
  \label{fig:fashion_mnist_rba}
\end{subfigure}
\begin{subfigure}{.35\textwidth}
  \centering
  \includegraphics[width=\linewidth]{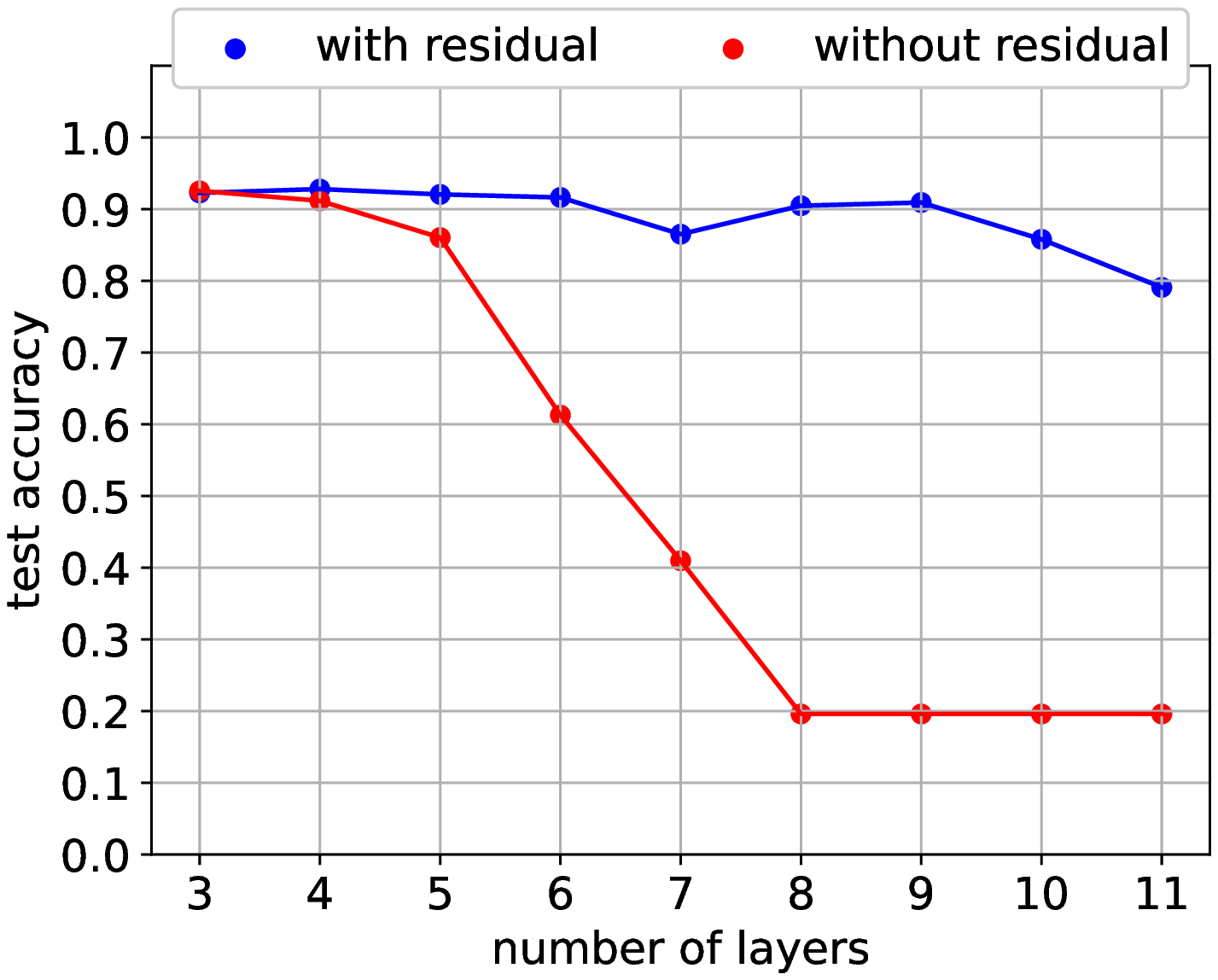}
  \caption{}
  \label{fig:svhn_rba}
\end{subfigure}
\begin{subfigure}{.35\textwidth}
  \centering
  \includegraphics[width=\linewidth]{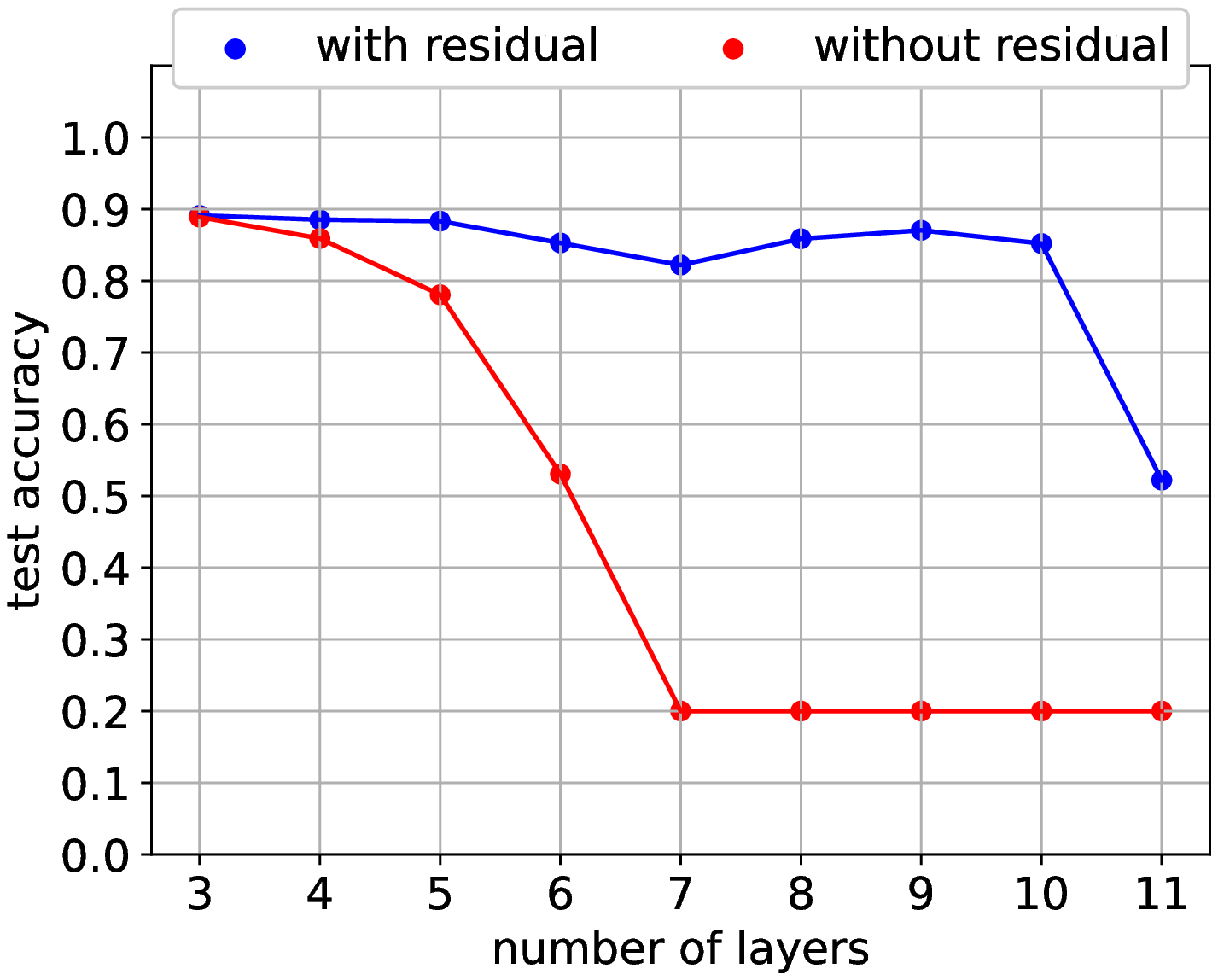}
  \caption{}
  \label{fig:norb_rba}
\end{subfigure}
}

% SDA
\noindent\makebox[\textwidth]{%
\begin{subfigure}{.35\textwidth}
  \centering
  \includegraphics[width=\linewidth]{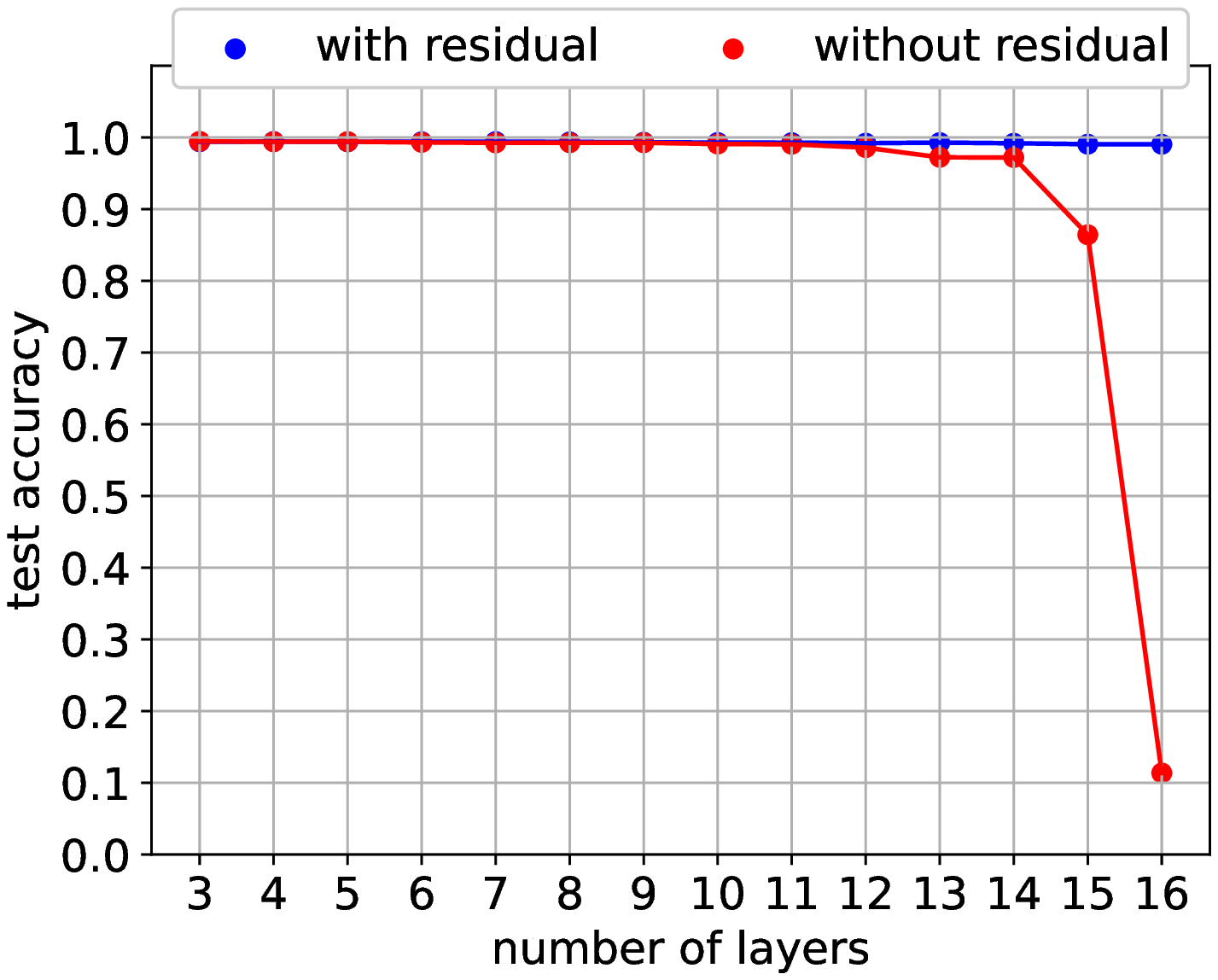}
  \caption{}
  \label{fig:mnist_sda}
\end{subfigure}%
\begin{subfigure}{.35\textwidth}
  \centering
  \includegraphics[width=\linewidth]{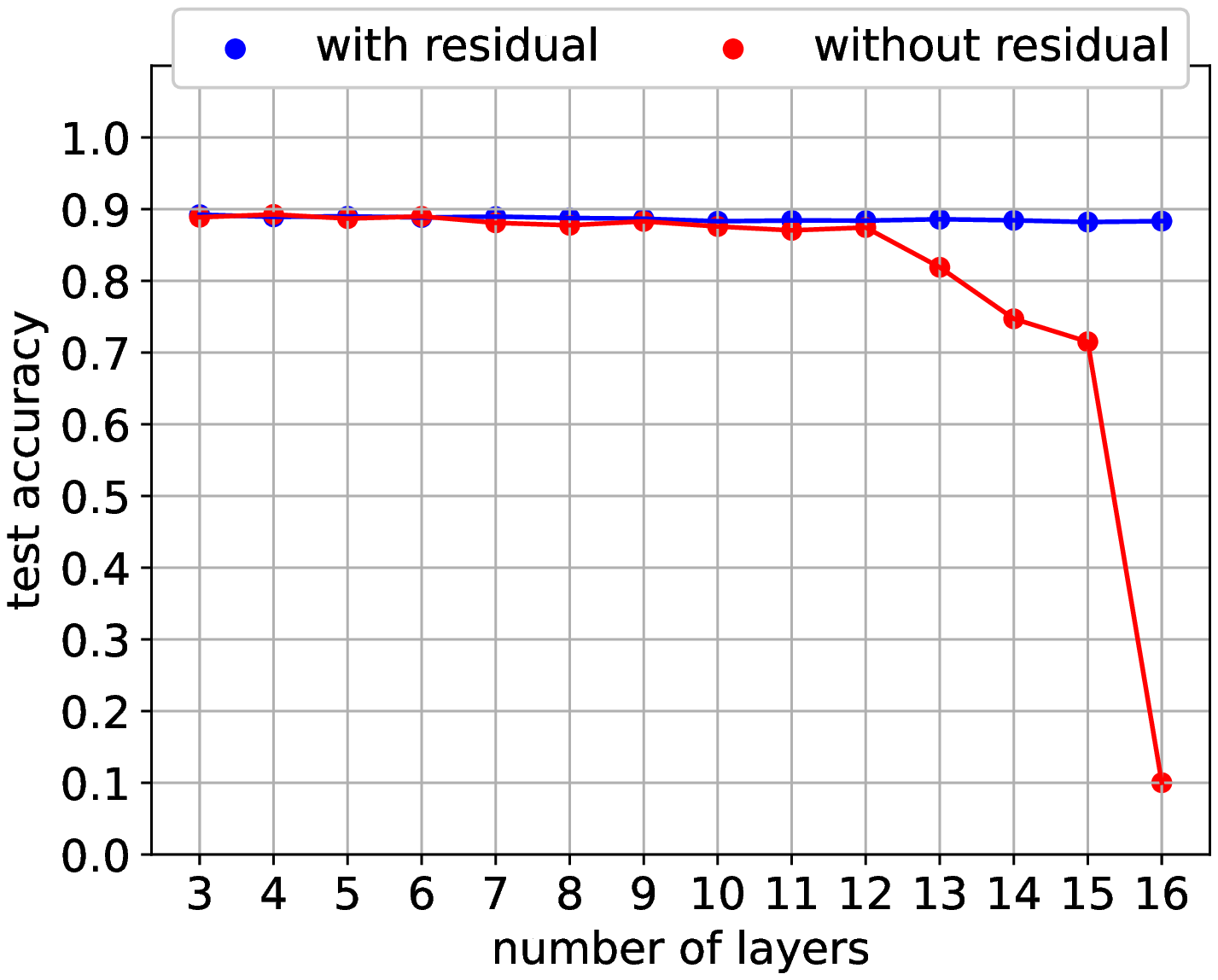}
  \caption{}
  \label{fig:fashion_mnist_sda}
\end{subfigure}
\begin{subfigure}{.35\textwidth}
  \centering
  \includegraphics[width=\linewidth]{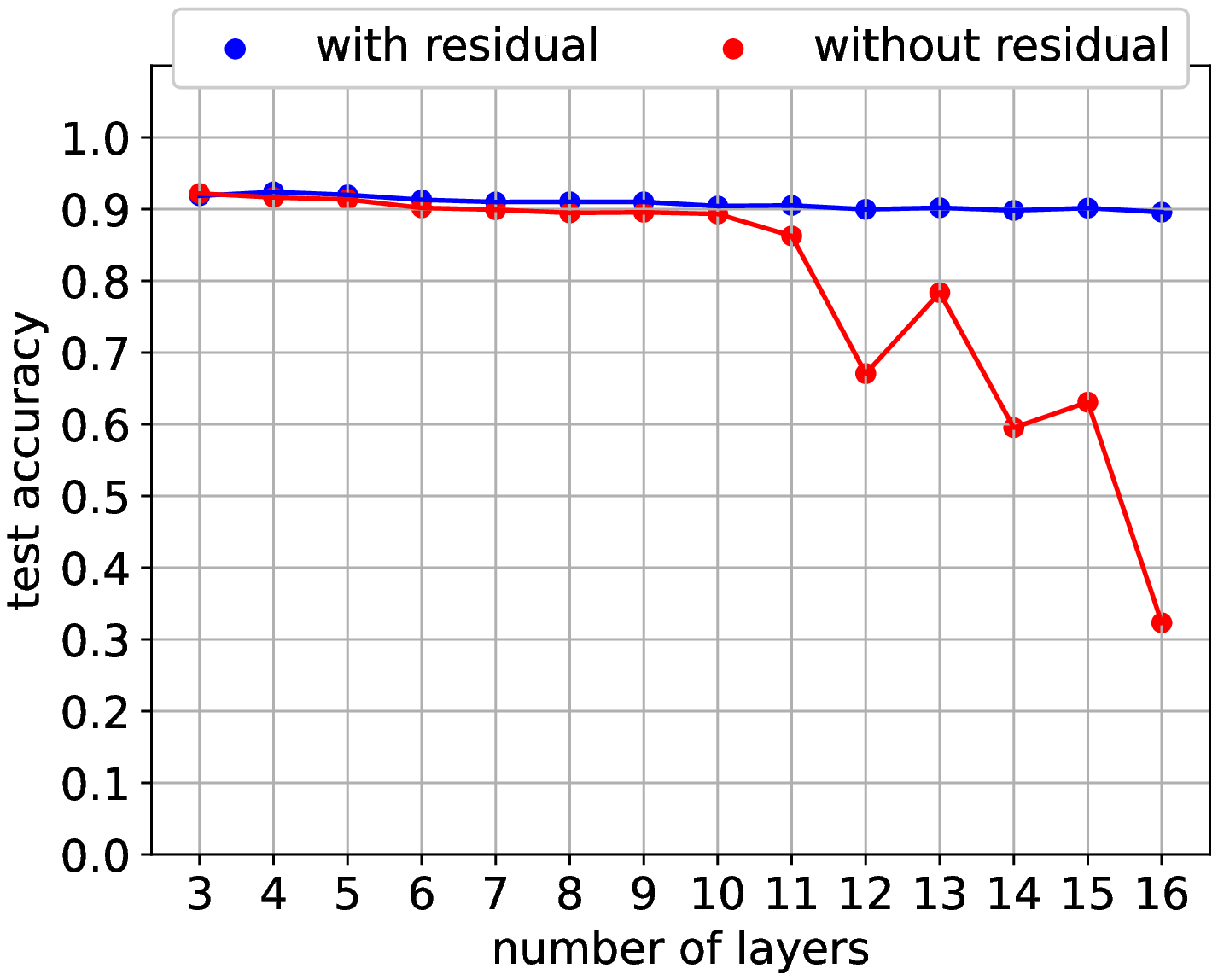}
  \caption{}
  \label{fig:svhn_sda}
\end{subfigure}%
\begin{subfigure}{.35\textwidth}
  \centering
  \includegraphics[width=\linewidth]{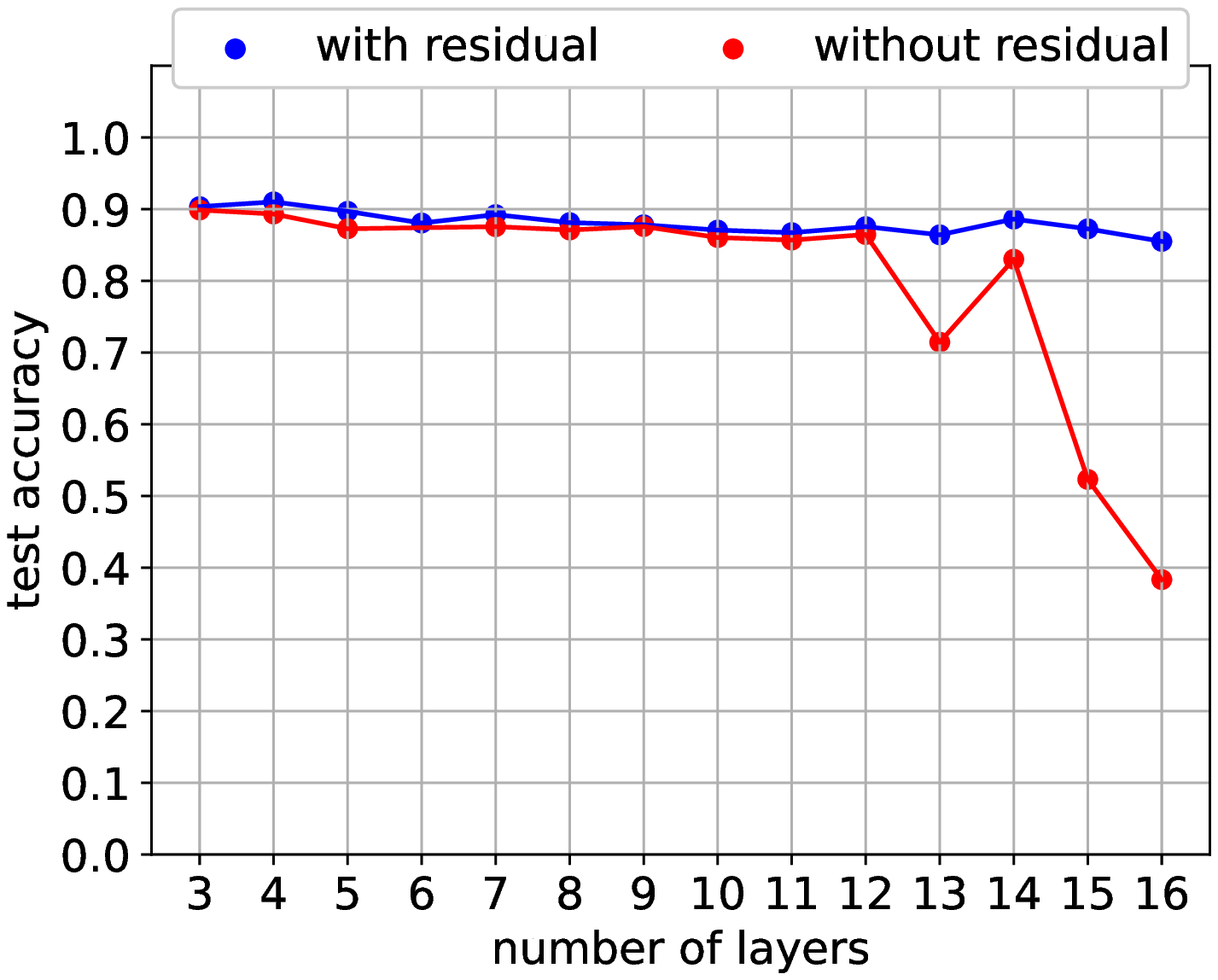}
  \caption{}
  \label{fig:norb_sda}
\end{subfigure}
}

% EM
\noindent\makebox[\textwidth]{%
\begin{subfigure}{.35\textwidth}
  \centering
  \includegraphics[width=\linewidth]{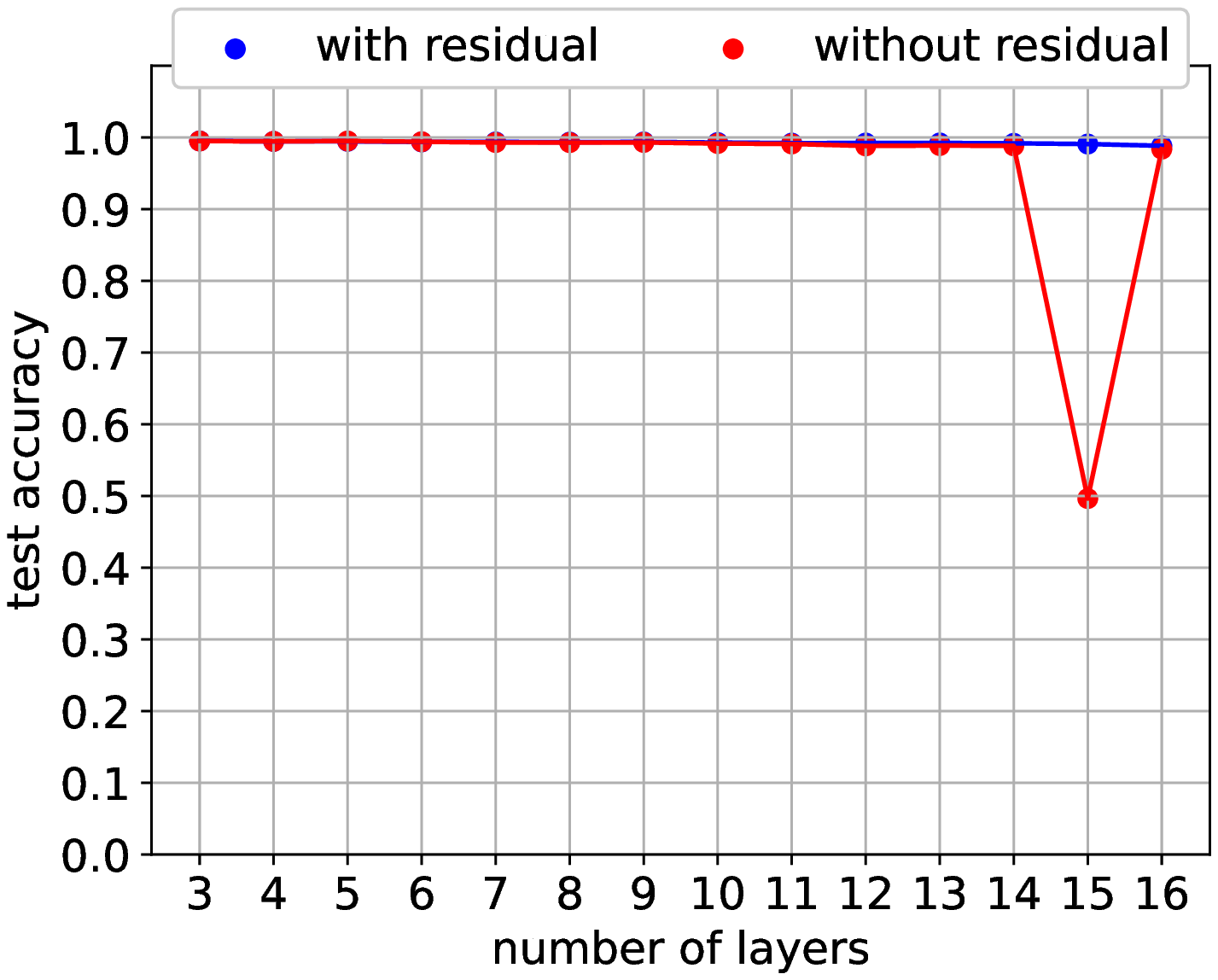}
  \caption{}
  \label{fig:mnist_em}
\end{subfigure}
\begin{subfigure}{.35\textwidth}
  \centering
  \includegraphics[width=\linewidth]{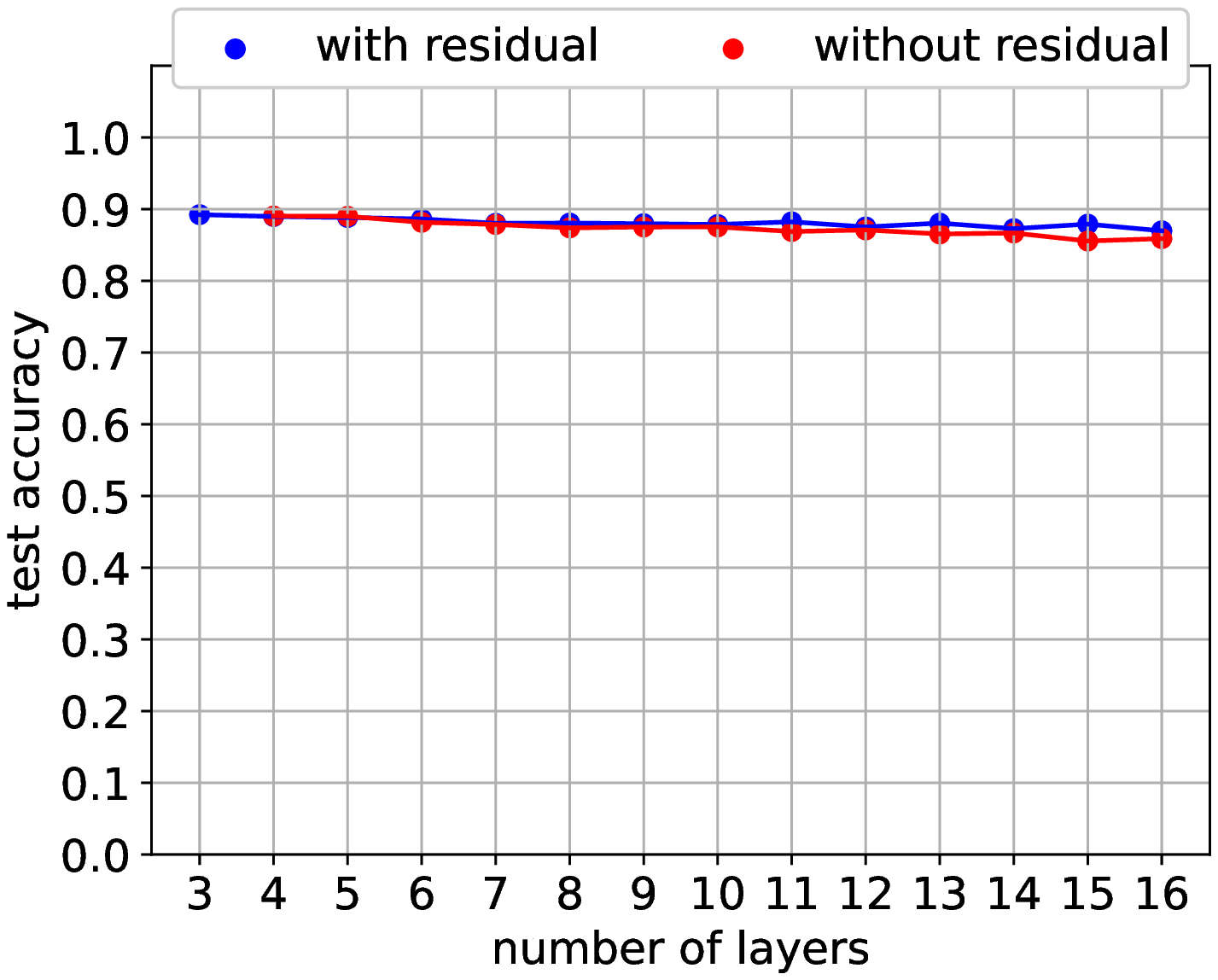}
  \caption{}
  \label{fig:fashion_mnist_em}
\end{subfigure}
\begin{subfigure}{.35\textwidth}
  \centering
  \includegraphics[width=\linewidth]{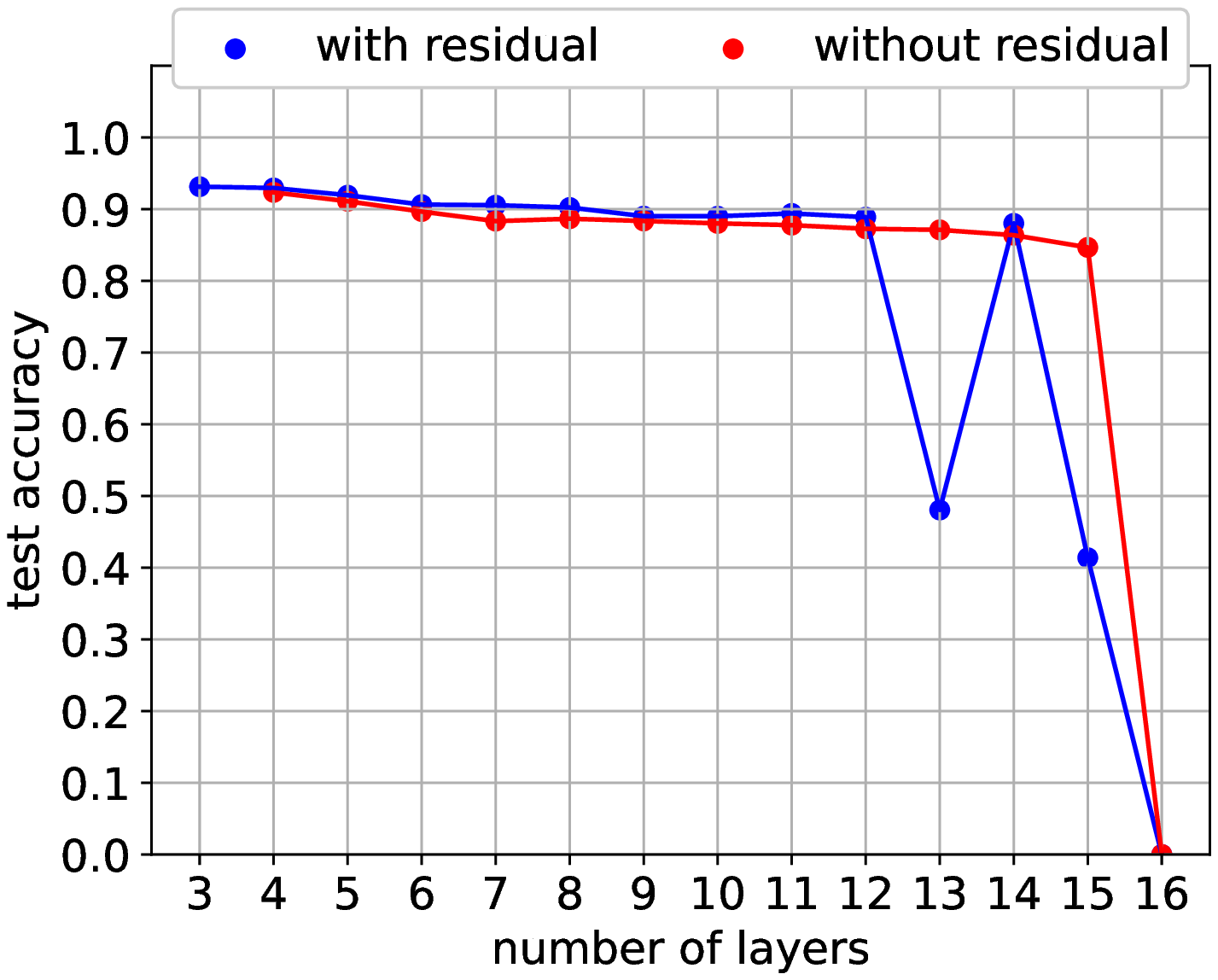}
  \caption{}
  \label{fig:svhn_em}
\end{subfigure}
\begin{subfigure}{.35\textwidth}
  \centering
  \includegraphics[width=\linewidth]{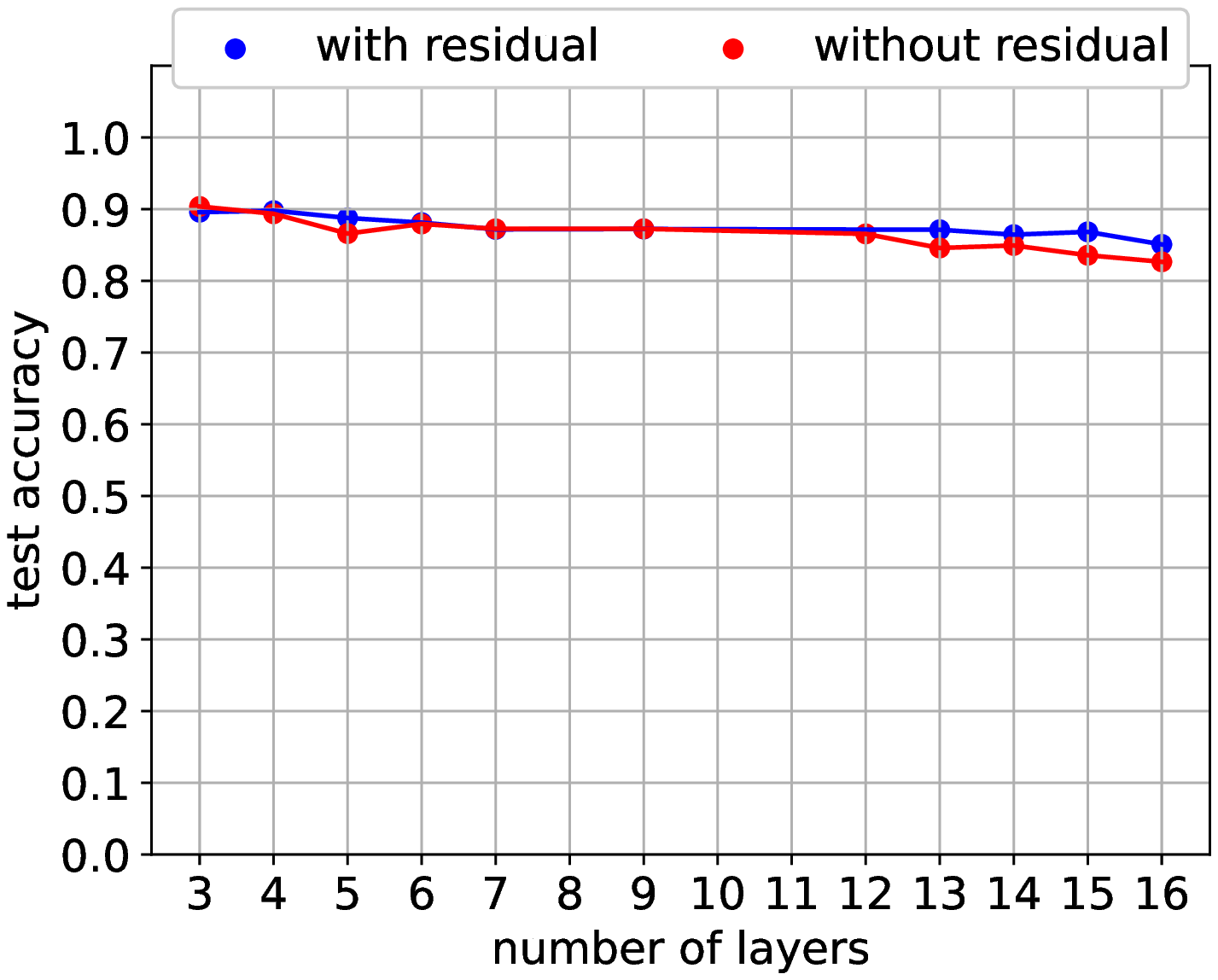}
  \caption{}
  \label{fig:norb_em}
\end{subfigure}
}
\caption{Results for capsules networks at different depths, trained with (blue curve) and without (red curve) the use of skip connections. Plotted are the accuracies using RBA (first row), SDA-routing (second row) and EM-routing (third row) for MNIST (\ref{fig:mnist_rba}, \ref{fig:mnist_sda} and \ref{fig:mnist_em} respectively), Fashion-MNIST (\ref{fig:fashion_mnist_rba}, \ref{fig:fashion_mnist_sda} and \ref{fig:fashion_mnist_em} respectively), SVNH (\ref{fig:svhn_rba}, \ref{fig:svhn_rba} and \ref{fig:svhn_em} respectively), and Small-NORB (\ref{fig:norb_rba}, \ref{fig:norb_sda} and \ref{fig:norb_em} respectively).}
\label{fig:routing_plots}
\end{figure}

We can better analyze the training behavior at different depths in figure \ref{fig:routing_plots} with skip (blue line) and without skip connections (red line). The top row (\ref{fig:routing_plots}a-d) shows the trend for RBA, the second row (\ref{fig:routing_plots}e-f) for SDA, last row corresponds to EM-routing (\ref{fig:routing_plots}g-j). We show all three routing algorithms for MNIST (first column), Fashion-MNIST (second column), SVHN (fourth column) and Small-NORB (fourth column). From this figure we can extract that in the case of RBA, after $5$ to $6$ layers the performance of the network drops without using skip connections and above $7$ or $8$ layers, performance  drops to around chance. On the other hand, with skip connections the performance keeps stable and the drop in performance happens much later.

In the case of SDA-routing, the benefits of using residual connections between capsule layers appears later, but after layer $13-14$ the network with residual connections performs significantly better on every dataset. Finally, for EM-routing, we can observe positive impact of residual connections on MNIST Fashion-MNIST and Small-NORB. However, there were two cases on the SVHN dataset (with $13$ and $15$ layers) where surprisingly, the network with residual connections performed slightly worse.

The results of our extensive experimental evaluation shows that there is an improvement in the performance of deep capsule networks using residual connections, this improvement was quite significant for the case of RBA. For SDA-routing and EM-routing, our results show that both routing strategies also benefited from skip connections, although to a lesser degree than for RBA. 

%
% CONCLUSION
%
\section{Conclusions \& Future Work}

In related work residual connections are either used before the capsule layers in the classical convolutional part or only a single capsule layer is used \cite{ai2021rescaps,rajasegaran2019deepcaps}. In this paper we have shown that its indeed possible to use residual connections together with multiple capsule layers. More precisely, we showed experimentally that training deep capsule networks greatly benefit from residual connections in terms of performance and stability. We experimented with three different routing algorithms on four datasets, and were able to train deep capsule networks in all cases. Even so, the test accuracy for deeper networks is significant larger when using residuals in almost all configurations. Therefore, we believe that in future work an extension of this work to convolutional capsule layers would further improve the accuracy, reduce the computational complexity and also enable the training on more complex datasets. Additionally, through the use of auto-tuning for deep capsule networks \cite{peer2021autotuning}, we could remove unused capsule layers from trained residual capsule networks and further improve the performance.

%
% BIBLIOGRAPHY
%
\bibliographystyle{splncs04}
\bibliography{mybib}

\end{document}